\documentclass[a4paper, 10pt, conference]{ieeeconf}      

\IEEEoverridecommandlockouts                              

\usepackage{times}

\usepackage{multicol}

\usepackage{epsfig}
\usepackage{graphicx}
\usepackage{amsmath}
\DeclareMathOperator{\arctantwo}{arctan2}
\usepackage{amssymb}
\usepackage{graphics} 
\usepackage{epsfig} 
\usepackage{mathptmx} 

\usepackage{url}
\usepackage{booktabs} 
\usepackage[table]{xcolor}
\usepackage{multirow}
\usepackage{float}
\usepackage{placeins}
\usepackage{graphicx} 
\usepackage[font=small]{caption}
\usepackage{subcaption}
\usepackage{siunitx}
\usepackage{etoolbox}
\robustify\bfseries

\usepackage{multicol}
\usepackage[ruled, shortend, linesnumbered]{algorithm2e}

\usepackage{breqn}  
\usepackage[pagebackref=true,breaklinks=true,colorlinks,bookmarks=false]{hyperref}

\newcommand*\blue{\color{black}}

\title{
Stein ICP for Uncertainty Estimation in Point Cloud Matching
}

\author{Fahira Afzal Maken$^{1,*}$, Fabio Ramos$^{1,2}$ and Lionel Ott$^{3}$
\thanks{$^*$ Corresponding author: fafz3958@uni.sydney.edu.au}
	\thanks{$^1$ School of Computer Science, The University of Sydney, Australia}
	\thanks{$^2$ NVIDIA, USA}
	\thanks{$^3$ ETH Z\"urich, Switzerland}%
}

\begin{document}

\maketitle
\thispagestyle{empty}
\pagestyle{empty}

\begin{abstract}
   Quantification of uncertainty in point cloud matching is critical in many tasks such as pose estimation, sensor fusion, and grasping. Iterative closest point (ICP) is a commonly used pose estimation algorithm which provides a point estimate of the transformation between two point clouds. There are many sources of uncertainty in this process that may arise due to sensor noise, ambiguous environment, initial condition, and occlusion. However, for safety critical problems such as autonomous driving, a point estimate of the pose transformation is not sufficient as it does not provide information about the multiple solutions. Current probabilistic ICP methods usually do not capture all sources of uncertainty and may provide unreliable transformation estimates which can have a detrimental effect in state estimation or decision making tasks that use this information. In this work we propose a new algorithm to align two point clouds that can precisely estimate the uncertainty of ICP's transformation parameters. We develop a Stein variational inference framework with gradient based optimization of ICP's cost function. The method provides a non-parametric estimate of the transformation, can model complex multi-modal distributions, and can be effectively parallelized on a GPU. Experiments using 3D kinect data as well as sparse indoor/outdoor LiDAR data show that our method is capable of efficiently producing accurate pose uncertainty estimates.
   
\end{abstract}

\IEEEpeerreviewmaketitle

\section{Introduction}
 Point cloud registration plays a fundamental role in many robotics and computer vision tasks such as localization and mapping \cite{acm}, autonomous navigation \cite{autnavShi2017A6N}, pose estimation \cite{fastgpupark2010fast}, surgical guidance \cite{micca}, and augmented reality \cite{augment} to name a few. Iterative closest point (ICP) \cite{besl_1992} is the gold standard registration algorithm which estimates a relative transformation between two point clouds. Given an initial estimate, ICP minimizes the Euclidean distance between pairs of matching points from both point clouds in an iterative manner. 

\begin{figure}[bt]
    \centering
\begin{subfigure}{0.48\textwidth}
  
    \includegraphics[page=1,width=0.18\textwidth]{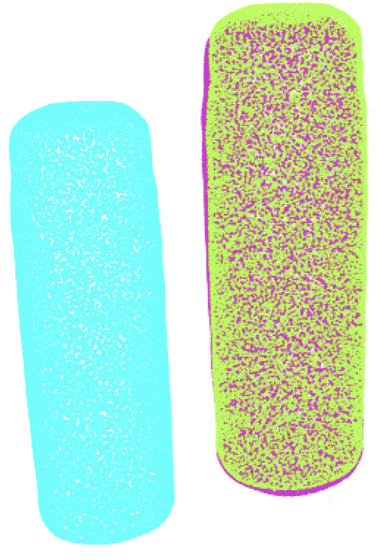}
      \hfill
    \includegraphics[page=1,width=0.61\textwidth]{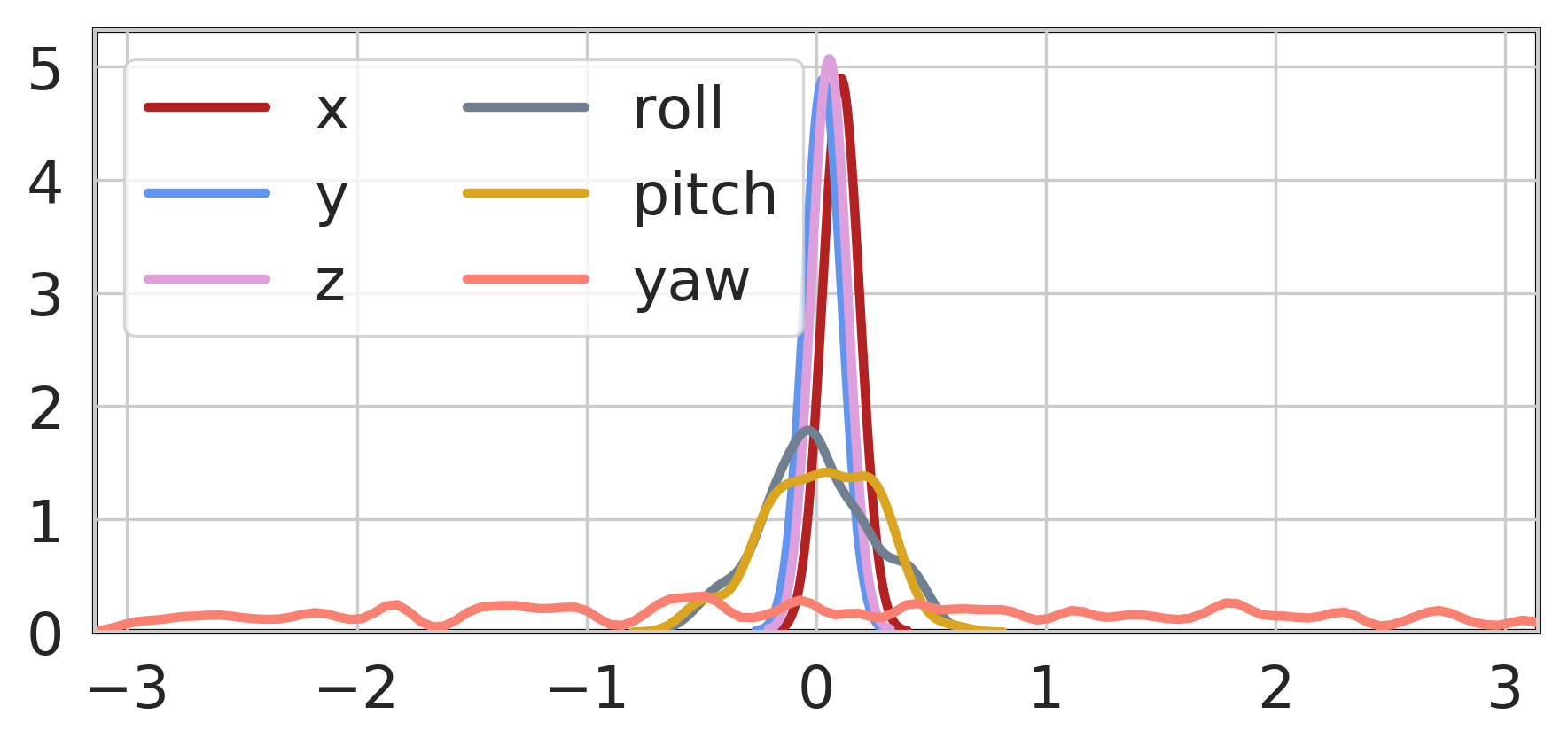}
    
\end{subfigure}

\begin{subfigure}{0.48\textwidth}
     \includegraphics[width=0.37\textwidth]{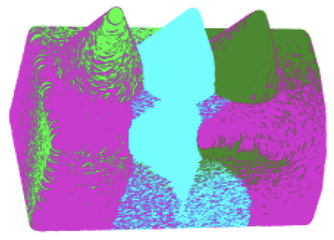}
      \hfill
    \includegraphics[page=1,width=0.61\textwidth]{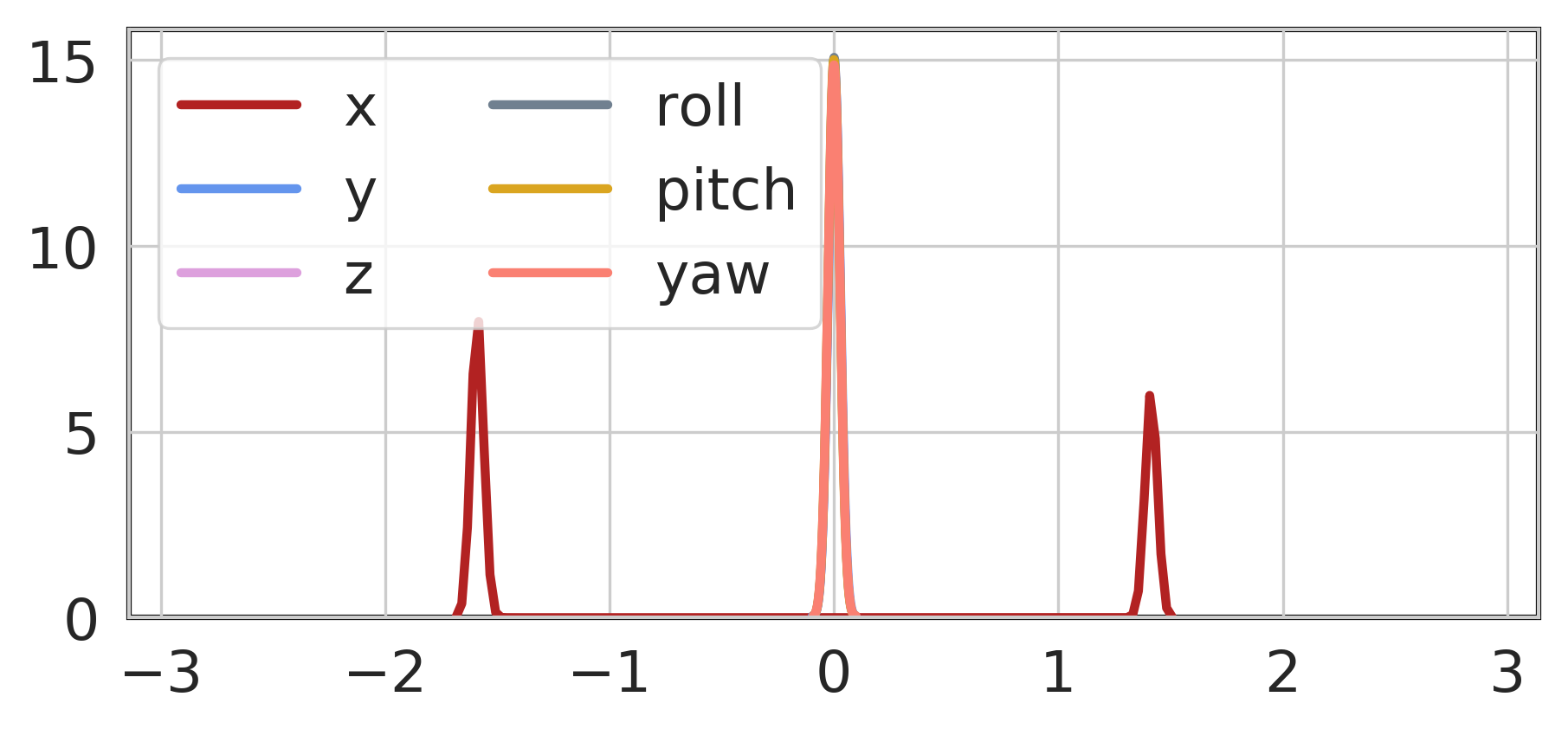}
    
\end{subfigure}

    \setlength{\belowcaptionskip}{-15pt}
    \caption{
    Our method estimates ICP pose distributions (right) for the solution (green) when a source cloud (cyan) is aligned to a reference cloud (magenta). Flat yaw distribution  (top right) characterizes the rotational ambiguity of the shape. For the bottom figure, our method estimates two possible modes for x as shown in the corresponding kernel density estimate on the right which a deterministic ICP algorithm cannot capture.
    }
    \label{fig:persuasive_context}
\end{figure}

The ICP pose alignment process is adversely affected by different sources of error and uncertainty.
These include initial pose uncertainty, sensor noise, partial overlap, multiple local minima of the cost function, and under-constrained or ill-posed cases, \eg long featureless corridors or rotational symmetric objects such as bottles that admit infinite solutions. In the alignment of an under-constrained object where there is ambiguity in the matching, uncertainty relates to the geometry of the object. This type of uncertainty is intrinsic to the problem and known as {\em epistemic}. For example, in Figure \ref{fig:persuasive_context} (top), the broad yaw distribution captures the rotational symmetry of the bottle. For a rectangular building block (bottom left), source cloud (cyan) can match to any of the two sides of the reference cloud (magenta), thus giving bi-modal $x$ distribution (right) which a deterministic ICP algorithm cannot capture. 

The quantification of the transformation uncertainty is crucial for many other tasks, especially those requiring robust pose estimates of an autonomous vehicle in an urban environment. In order to accurately localize a vehicle, a single source of pose information is usually not enough. Pose information from different sensors are usually probabilistically fused together \cite{uncertainCaglioti1994ImprovingPE,odometTur2007OntoCT} in a state estimation framework such as simultaneous localization and mapping (SLAM) \cite{Barczyk2017}. In such a framework, ICP pose estimates are fused with odometry measurements or GPS observations using Extended Kalman Filter (EKF) \cite{ekf1, ekf2}, particle filter \cite{pf} or GraphSLAM \cite{gslam} which require information about the uncertainty of pose parameters. This uncertainty estimate encompasses information of the reliability of sensory data and aids in improving the estimates of the pose parameters. In the localization and mapping tasks, this uncertainty information can improve loop closure detection and yield better maps. In the task of scene reconstruction, the amount of uncertainty in the transformation parameters can indicate a tangential drift that may occur in the registration of two plane objects. 

Bayesian methods \cite{Neal1996} can be used to incorporate uncertainty into ICP pose estimates. These methods provide a probabilistic framework 
in which Bayes' rule is used to obtain a posterior distribution given a prior distribution and a likelihood function. However, Bayesian methods tend to be computationally intractable for complicated likelihood functions and high-dimensional problems. To address the tractability issue, modern Bayesian methods rely on approximate techniques such as Markov Chain Monte Carlo (MCMC)~\cite{Neal2010} and variational inference (VI)~\cite{advancevi} which can scale to large problems using stochastic gradient descent (SGD)~\cite{robbins1951}.

MCMC methods approximate the intractable posterior distribution by drawing samples from the prior and likelihood functions. In contrast, VI transforms the problem into an optimization process that reduces the Kullback-Leibler (KL) divergence between the intractable posterior distribution and an analytically tractable variational distribution. MCMC will converge to the underlying distribution but can be slow in practice. VI on the other hand is typically faster but the expressiveness of the variational distribution is restricted and may not capture the complexity of the true posterior distribution.

In this paper we formulate a Stein variational gradient descent (SVGD) \cite{steinNIPS2016_6338} method to approximate the posterior distribution of ICP pose parameters. SVGD combines the accuracy and flexibility of MCMC with the speed of VI using  gradient descent. SVGD approximates the intractable posterior distribution with a non-parametric representation given by a set of particles. These particles are optimized with a functional gradient descent of KL divergence and provide a reliable uncertainty estimation by incorporating a {\em repulsive} term that prevents particles from clustering together and allows them to capture a wide range of complex distributions. 
These particles can be updated jointly and benefit from GPU parallelization. .

\noindent\textbf{Contribution:} The main contribution of this paper is a non-parametric point cloud registration method based on Stein variational gradient descent to estimate ICP's pose uncertainty. Our method exploits GPU-computation for increased speed and captures both epistemic and aleatory uncertainty \cite{aleaun} intrinsic in the matching problems. {\blue A python implementation of the method is publicly available\footnote{\url{https://bitbucket.org/fafz/stein-icp/src/master/}}}.

\section{Related Work}
There exists a large number of point cloud registration methods as reviewed in \cite{Pomerleau2015, Rusinkiewicz2001}. They can be broadly categorized as algorithms that provide point estimates and probabilistic outputs.

\subsection{Point-Based ICP Algorithms}
 Point estimate registration algorithms produce a single estimate of the transformation between two point clouds. These algorithms mostly vary in how they select points to perform data association, which error metric they use, and what optimization technique they employ for their cost function \cite{Rusinkiewicz2001}. 
 
 ICP algorithms which vary in the selection of points include those using random sampling \cite{Masuda1995} and uniform sub-sampling \cite{Masuda2001}. Other ICP methods have different cost functions. Notable choices are point-to-point \cite{besl_1992} point-to-plane \cite{Chen1992} and plane-to-plane \cite{Segal2009b}. The ICP cost function can be optimized using either closed form solutions \eg singular value decomposition (SVD) \cite{svd66357613edf64c9e97ef7812714116c2}, and quaternions \cite{Horn87closed-formsolution} or other methods \eg Levenberg-Marquardt~\cite{Fitzgibbon2003}, simulated annealing \cite{Luck2000}, and stochastic gradient descent \cite{fahira2018}.

Another line of research deals with eliminating outliers due to partial overlap, sensor noise and other complexities. This includes the use of a trimmed square cost function to estimate the optimal transformation \cite{Chetverikov2002}, and rejecting pairs with distance larger than a certain threshold \cite{Segal2009b, Masuda1995}.

An alternative class of registration technique reformulates ICP as an alignment of two Gaussian mixtures where a statistical discrepancy measure is minimized between the two mixtures~\cite{gmm2Eckart2018HGMRHG}. In this case Expectation Maximization \cite{EmalgoDempster77maximumlikelihood} can be used to provide robust solutions \ie \cite{emfilter8954154}. Learning-based point cloud registration methods employ neural networks to provide pose parameters \cite{PointNetLKR,deepclWang2019DeepCP}. These methods rely on the data used during training and may not necessarily generalize well to an unseen dataset.

\subsection{Uncertainty-Based ICP Algorithms}
This class of ICP provides uncertainty estimates for the transformation parameters. The uncertainty of the transformation can be estimated using closed form solutions \cite{Biber2003,Bengtsson2003,Bosse2008,Censi2007}, and sampling-based approaches \cite{Bengtsson2003,Iversen2017a,Maken2020EstimatingMU}. Closed form solutions require the Hessian of the cost function. Among closed form solutions such as in \cite{Biber2003,Bengtsson2003,Bosse2008}, errors due to sensor noise are not properly captured as pointed out in \cite{Censi2007,Prakhya}. \cite{Censi2007} considers the effect of sensor noise in the covariance structure but  \cite{ral,Mendes2016ICPbasedPS} indicate that it is overoptimistic and hence not suitable for sensor fusion. Closed-form solutions are efficient but do not capture the uncertainty of the data association into account in the covariance estimation.

Sampling based approaches such as \cite{Bengtsson2003,Iversen2017a} provide high quality samples of pose parameters but are computationally expensive. A recently proposed scalable probabilistic ICP method, Bayesian ICP \cite{Maken2020EstimatingMU}, employs stochastic gradient Langevin dynamics (SGLD)~\cite{Welling} to provide samples of the posterior distribution in an online manner. However, due to it still being an MCMC style method it suffers from the inherent limitations and challenges of parallelizing any MCMC method. 

Other recently proposed methods such as \cite{ral,celloDBLP:journals/corr/abs-1810-01470} take initial pose estimates into account in the pose uncertainty prediction. \cite{ral} combines a closed-form solution to incorporate sensor noise with the unscented transform \cite{unscent847726} to capture other sources of uncertainty. This method relies on an accurate covariance of the initial pose uncertainty to produce reliable uncertainty estimation which is not always available. 

Our method can take prior information into account to estimate the posterior distribution over transformation parameters. When a prior is available, our method leverages this information to achieve faster and more accurate convergence. 
In comparison to Bayesian ICP which uses MCMC sampling, our method employs Stein particles that can be propagated in parallel through Stein's gradients, directly modeling the interactions between the particles and promoting diversity in the solution. This is particularly important in multi-modal problems as we demonstrate in a few examples. Additionally, our method is directly amenable to parallelization, therefore suitable to GPU computation. 

\section{Preliminaries}
In this section we first give an overview of standard ICP \cite{besl_1992}, followed by a description of stochastic gradient descent ICP (SGD-ICP) \cite{fahira2018} which constitutes a core component of our method. 

\subsection{Standard ICP}
Standard ICP aligns two input point clouds by iteratively performing the following two steps until convergence:
\begin{enumerate}
     \item Establish the point pairings on the basis of minimum Euclidean distance using an initial guess $\boldsymbol{\theta} = \{x, y, z, \text{roll}, \text{pitch}, \text{yaw}\}$. The commonly used point-to-point distance metric to select the corresponding pair of points is expressed as follows:
\begin{equation}
    \text{point-to-point}(\mathbf{s_i}', \mathcal R) = \min_{\mathbf{r} \in \mathcal R} \left\Vert \mathbf{r} - \mathbf{s_i}' \right\Vert,
    \label{eq:correspondences}
\end{equation}
where  $\mathbf{ s_i}, \mathbf {r_i} \in \mathbb R^3$ are points in 3D space belonging to a source cloud $\mathcal S =  \{\mathbf{s_i}\}_{i =1}^{N}$ and a reference cloud $\mathcal R = \{\mathbf{r_i}\}_{i =1}^{M}$ respectively. $\mathbf{s_i}' =(R \; \mathbf{s_i}+ \mathbf{u})$ is a transformed point in the source cloud, $\mathbf{u} \in \mathbb R^{3 \times 1}$ is a translation vector comprising of $\boldsymbol{\theta}_{1:3}$ and $R \in \mathbb R^{3 \times 3}$ is a rotation matrix parametrized by $\boldsymbol{\theta}_{4:6}$.
    \item Find the transformation that minimizes a loss function defined by the Euclidean distance between each pair of corresponding points. The point-to-point cost function has the following form:
    \begin{equation}
    \operatorname*{argmin}_{\boldsymbol{\theta}}  \mathcal L(\boldsymbol{\theta}) = \frac{1}{N}\sum_i^N  ||(R \; \mathbf{s_i}+ \mathbf{u}) - \mathbf{r_i}||^2.
    \label{eq:icp-objective1}
\end{equation}
\end{enumerate}
\subsection{Stochastic Gradient Descent ICP}\label{sgdicp}
SGD-ICP \cite{fahira2018} optimizes the ICP cost function \eqref{eq:icp-objective1} using stochastic gradient descent (SGD) \cite{robbins1951}. Instead of using all available points, SGD-ICP samples a mini-batch $\mathcal M$ of $m$ points from the source cloud to find corresponding points from the reference cloud. SGD-ICP computes gradients of the cost function \eqref{eq:icp-objective1} to update the transformation parameters $\theta$ with the following update rule:
\begin{equation}
    \boldsymbol{\theta}^{t+1} = \boldsymbol{\theta}^{t} - \eta A \bar{g} (\boldsymbol{\theta}^{t}, \mathcal{M}^{t}),
    \label{eq:sgd-fomulation_simplified}
\end{equation}
where $\boldsymbol{\theta}^{t}$ and $\boldsymbol{\theta}^{t+1}$ are the values of the pose parameters at the current and next iteration respectively. The matrix $A \in \mathbb R^{6 \times 6}$ acts as a pre-conditioner for the gradients and the learning rate $\eta$ dictates how quickly parameter values change. The average gradients $\bar g$ of the parameter vector $\boldsymbol{\theta}$ with respect to the loss $\mathcal L$ are computed as follows:
\begin{align}
    \bar{g} (\boldsymbol{\theta}^t_{1:3}, \mathcal{M}^t) & = \frac{1}{m} \sum_i^m \Big((R^{t} \; \mathbf{s_i}+\mathbf{u}^t) - \mathbf{r_i} \Big) \frac{\partial \mathbf{u}^{t}}{\partial\boldsymbol{\theta}_{1:3}^t},
    \label{eq:trans_gradient}
    \\
    \bar{g} (\boldsymbol{\theta}^t_{4:6}, \mathcal{M}^t) & =  \frac{1}{m} \sum_i^m\Big(( R^{t} \; \mathbf{s_i}+\mathbf{u}^{t}) - \mathbf{r_i}\Big) \frac{\partial R^{t}}{\partial \boldsymbol{\theta}_{4:6}^{t}} \; \mathbf{s_i},
    \label{eq:rotation_gradient}
\end{align}
where $\bar g(\boldsymbol{\theta}^t_{1:3}, \mathcal M^t)$ are the gradients of the translation components and $\bar g(\boldsymbol{\theta}^t_{4:6}, \mathcal M^t)$ are the rotational gradients. 

The quality of the estimated pose of SGD-ICP is the same as that of standard ICP but it is computationally more efficient due to the mini-batch formulation \cite{fahira2018}. In the next section we employ the mini-batch gradient-based formulation of SGD-ICP within Stein variational inference to derive Stein ICP.


\section{Stein ICP}

Our proposed method, Stein ICP, makes the connection between the gradient-based stochastic optimization of {\blue an} ICP cost function and Stein variational inference. Specifically, Stein ICP utilizes the SGD-ICP \cite{fahira2018} gradients within the Stein variational gradient descent (SVGD) framework \cite{steinNIPS2016_6338}.
SVGD approximates an intractable but differentiable posterior distribution $p(\mathbf{x})$ by constructing a non-parametric variational distribution represented by a set of $K$ particles $\{\mathbf{x}_j\}_{j=0}^K$. These particles are updated iteratively by an update rule of the following form:
\begin{equation}
    \mathbf{x}_j  \gets  \mathbf{x}_j +  \eta \boldsymbol{\phi} (\mathbf{x}_j) \quad \forall j = 1, \ldots, K,  
    \label{eq:stein_transport}
\end{equation} 
where $\eta$ is a small step size, and 
$\boldsymbol{\phi} \colon \mathbb R^d \to \mathbb R^d$ characterizes a
perturbation direction {\blue that gives the steepest descent. 
$\boldsymbol{\phi}$ } should be chosen to push-forward the particle distribution closer to the target. The optimal value of the perturbation direction $\boldsymbol{\phi}$ can be obtained from the following expression:  
\begin{equation}
\boldsymbol{\phi}^* = \operatorname*{argmax}_{\boldsymbol{\phi} \in \mathcal{B} }  \bigg(  -   \frac{d}{d\eta} \text KL(q_{[\eta\boldsymbol{\phi}]} ~|| ~ p) \big |_{\eta = 0}  \bigg), 
\label{eq:stein_phistar_with_kl}
\end{equation}
where {\blue $q_{[\eta\boldsymbol{\phi}]}$ is specified nonparametrically by the updated particles, $\mathbf{x}^\prime = \mathbf{x} + \eta \boldsymbol{\phi}(\mathbf{x})$.} 
$\boldsymbol{\phi}^*$ is selected from 
{\blue a function set,} $\mathcal{B}${\blue,} to minimize the KL divergence between the particle distribution and the target \cite{matrixstein}. 

To obtain tractable and flexible solutions, SVGD chooses $\mathcal{B}$ to be in the unit ball of a vector-valued reproducing kernel Hilbert space (RKHS) $\mathcal{H}{\blue^d} = \mathcal{H} \times \cdots \times \mathcal{H}$,
where $\mathcal{H}^d$ is an RKHS formed by scalar-valued functions associated with a positive definite kernel $k(\mathbf{x},\mathbf{x}')$, that is, 
$\mathcal{B} = \{\boldsymbol{\phi} \in \mathcal{H}{\blue^d} \colon ||\boldsymbol{\phi}||_{\mathcal{H}{\blue^d} }\leq 1 \}.$ {\blue Formally, an RKHS is defined as a Hilbert space $\mathcal{H}{\blue^d}$  with inner product ${\langle\cdot,\cdot\rangle}_{\mathcal{H}{\blue^d}}$ and norm $||.||_{\mathcal{H}{\blue^d}}$ satisfying the reproducing property, $\forall f \in \mathcal{H}{\blue^d}, f(\mathbf{x}) = {\langle f,k(\mathbf{x},.)\rangle}_{\mathcal{H}{\blue^d}}$}. {\blue  A key observation to solve Eq.~\ref{eq:stein_phistar_with_kl} is that the optimization is a linear functional of $\boldsymbol\phi$ that connects to the Stein operator $\mathcal{A}$ in the following equation~\cite{steinNIPS2016_6338},}
\begin{equation}
-\frac{d}{d\eta} \text KL(q_{[\eta\boldsymbol{\phi}]} ~|| ~ p) \big |_{\eta = 0}  = \mathbb E_{\mathbf{x}\sim q}[\text trace(\mathcal{A}  \boldsymbol{\phi}(\mathbf{x}))],
\label{eq:operator_a}
\end{equation}
with
\begin{equation}
 \mathcal{A} \boldsymbol{\phi}(\mathbf{x})  = \nabla_{\mathbf{x}} \log p(\mathbf{x})  \boldsymbol{\phi} (\mathbf{x})^\top+ \nabla_{\mathbf{x}}\boldsymbol{\phi}(\mathbf{x}). 
\label{eq:stein_operator}
\end{equation}
{\blue By Stein's identity in Eq.~\ref{eq:operator_a}, we obtain 
$\frac{d}{d\eta} \text KL(q_{[\eta\boldsymbol{\phi}]} ~|| ~ p)=0$ when $p=q$}. Following from Eq. \ref{eq:stein_operator}, the gradient of the log posterior $\nabla_{\mathbf{x}} \log p(\mathbf{x}|D)$ can be expressed as:
\begin{equation}
  \nabla_{\mathbf{x}|D} \log p(\mathbf{x}) = \sum_{i=1}^{N} \nabla_{\mathbf{x}} \log p(d_i|\mathbf{x}) + \; \nabla_{\mathbf{x}} \log p(\mathbf{x}), 
  \label{eq:posterior}
\end{equation}
where $\nabla \log p(d_{i}|\mathbf{x})$ and $\nabla \log p(\mathbf{x})$ are the gradient of the log likelihood function, and the gradient of the log of the prior with respect to $\mathbf{x}$ respectively, for a dataset $D=\{d_i\}_{i=1}^N$ of $N$ data points. Note that the Stein operator does not depend on the normalization $p(D)$, also known as the marginal likelihood, which makes the computation of the posterior significantly easier. This makes SVGD a powerful tool for inference of intractable distributions. The Stein variational gradient descent algorithm follows from a closed form solution of \eqref{eq:stein_phistar_with_kl}, as shown in \cite{steinNIPS2016_6338}, and given by:
\begin{equation}
\boldsymbol{\phi}^*(\cdot)  = \mathbb E_{\mathbf{x}\sim q}[ \nabla_{\mathbf{x}} \log p(\mathbf{x}) k(\mathbf{x},\cdot) + \nabla_{\mathbf{x}} k(\mathbf{x},\cdot)].
\label{equ:phistar_closed_form}             
\end{equation}
In the above we omit the dependence on the data $D$ to simplify notation. Equation \eqref{equ:phistar_closed_form} provides the optimal update direction for the particles within $\mathcal H^d$.
In practice, a set of particles ${\{\mathbf{x}_j\}}_{j=1}^K$ approximates $q$ yielding the following update rule:
\begin{equation}
\mathbf{x}_j \gets \mathbf{x}_j  + \eta \hat {\boldsymbol{\phi}}^* (\mathbf{x}_j),
\label{stein_update}
\end{equation}
where $\hat {\boldsymbol{\phi}}^* (\mathbf{x}) $ is an approximate steepest direction given by
\begin{equation}
\hat {\boldsymbol{\phi}}^* (\mathbf{x}) =\sum_{j=1}^K [\nabla \log p(\mathbf{x}_j)k(\mathbf{x}_j, \mathbf{x}) +  \nabla_{\mathbf{x}_j}k(\mathbf{x}_j,\mathbf{x})]. 
\label{stein_phi_approx}
\end{equation}
The first gradient term in \eqref{stein_phi_approx} is weighted by a kernel function{\blue, which smooths the gradients,} and acts as the steepest direction for the log probability. The second term is the gradient of the kernel function. It can be viewed as a repulsive force which induces the spread among the particles and prevents them from collapsing to the local modes of the log probability.

In SGD-ICP $-\nabla \log p(D|\mathbf{x})$, {\blue in Eq.~\ref{eq:posterior},} is expressed as the gradients of the cost function in \eqref{eq:trans_gradient} and \eqref{eq:rotation_gradient}. $\nabla \log p(\mathbf{x})$ represents the gradient of the log of Gaussian priors for translations and von Mises priors for rotations \cite{Maken2020EstimatingMU}:
\begin{equation}
    \nabla_{\boldsymbol{\theta}} \log p(\boldsymbol{\theta}_{1:3})  = -(\boldsymbol{\theta}_{1:3} - \boldsymbol{\mu}_{1:3}) / \boldsymbol{\sigma}_{1:3},
    \label{trans_prior}
\end{equation} 
\begin{equation}
   \; \;\; \;\; \;\; \;\; \;\; \;\;   \nabla_{\boldsymbol{\theta}} \log p(\boldsymbol{\theta}_{4:6})  = -\boldsymbol{\kappa}_{4:6} \sin (\boldsymbol{\theta}_{4:6} - \boldsymbol{\mu}_{4:6}),
    \label{rot_prior}
\end{equation}
where $\boldsymbol{\mu}_{1:6}$ represents the mean while $\boldsymbol{\sigma}_{1:3}$, and $\boldsymbol{\sigma}_{4:6} = 1 / \boldsymbol{\kappa}_{4:6}$ represent the variance of the prior distributions of the translation and rotation components respectively.

In Stein ICP, we replace the gradients in \eqref{stein_phi_approx} with the mini-batch gradients of the ICP cost function shown in \eqref{eq:trans_gradient} and \eqref{eq:rotation_gradient}, along with the gradients of priors expressed in \eqref{trans_prior} and \eqref{rot_prior}. By doing this, Stein variational gradients are independently obtained for translations ($\boldsymbol{\theta}_{1:3}$) and rotations ($\boldsymbol{\theta}_{4:6}$), for all the  particles in \eqref{stein_icp_phi}, which are then updated using \eqref{stein_update} .
\begin{multline}
\hat {\boldsymbol{\phi}}^* (\boldsymbol{\theta}_{}) =\sum_{j=1}^K \Big[-\Big(N\bar{g} (\boldsymbol{\theta}_{{}_j}, \mathcal{M})  + \nabla_{\boldsymbol{\theta}} \log p(\boldsymbol{\theta}_{{}_j})  \Big ) k(\boldsymbol{\theta}_{{}_j}, \boldsymbol{\theta}_{{}}) \; \\+  \nabla_{\boldsymbol{\theta}}k(\boldsymbol{\theta}_{{}_j},\boldsymbol{\theta}_{{}})\Big]. 
\label{stein_icp_phi}
\end{multline}

In SVGD the kernel plays a critical role in weighing the gradients and dispersing the particles. In Stein ICP, for translation parameters we use an RBF kernel $k(\boldsymbol{\theta}_{1:3}, \boldsymbol{\theta}'_{1:3} ) = \exp (-\frac{1}{h}||\boldsymbol{\theta}_{1:3} -  \boldsymbol{\theta}'_{1:3}||^2_2)$, where $h$ is the bandwidth of the kernel. For the rotational parameters, we use a modified version of the RBF kernel as the Euclidean distance is not directly applicable to angles. In particular we use the following modified RBF kernel for rotations:
\begin{multline}
k(\boldsymbol{\theta}_{4:6}, \boldsymbol{\theta}'_{4:6}) = \exp \Bigg[-\frac{1}{h}\bigg (\arctantwo  \Big( \\  \sin(\boldsymbol{\theta}_{4:6}-\boldsymbol{\theta}'_{4:6}), \cos(\boldsymbol{\theta}_{4:6}-\boldsymbol{\theta}'_{4:6}) \Big ) \bigg)^2 \Bigg].
\end{multline}
In summary, Stein ICP begins by initializing a set of $K$ particles randomly. In general the implementation of Stein variational inference does not depend on the initial distribution \cite{steinNIPS2016_6338}. In practice, the particles can be initialized using prior knowledge, \eg readings from an inertial measurement unit or from a previous solution. Each particle represents the transformation of a mini-batch sampled from a source cloud $\mathcal {S}$ to a reference cloud $\mathcal {R}$, producing $K$ transformed mini-batches. 
Next for all points in each transformed mini-batch, corresponding closest points from a reference cloud are sought and stored in pairs using \eqref{eq:correspondences}. Then mean gradients are estimated for all the matching pairs belonging to each of the $K$ particles using \eqref{eq:trans_gradient} and \eqref{eq:rotation_gradient}. Next, Stein variational gradients are obtained independently for translations and rotations using \eqref{stein_icp_phi} which are then used to update each particle with \eqref{stein_update}. This procedure is repeated for $T$ iterations producing $K$ particles representing the posterior distribution. 

  
 \section{Experiments}
In the experiments we demonstrate the ability of Stein ICP to obtain high quality pose distributions while its formulation leads to significant computational gains when using a GPU.

To gain an intuitive insight into the behaviour of Stein ICP, we use objects with obvious distributions from an RGB-D dataset \cite{objectdata5980382}. For the quantitative evaluations we use a challenging point cloud alignment dataset \cite{laserdataPomerleau2012ChallengingDS} which contains different sequences recorded in both structured and unstructured environments. 
We provide comparisons with three other uncertainty aware ICP methods, namely Bayesian ICP \cite{Maken2020EstimatingMU}, Cov-3D ICP \cite{ral}, and Closed-form ICP \cite{Censi2007}. 

As there is no ground truth distribution available for the point cloud alignment we compute this ground truth information using Monte Carlo sampling. To obtain these Monte Carlo samples in a timely manner, we use SGD-ICP, which produces solutions equivalent to standard ICP methods in quality but much more quickly \cite{fahira2018}. We run $1000$ instances of SGD-ICP with initial transformation estimates sampled from $\pm$\SI{1}{\meter} and $\pm$\SI{0.1745}{\radian} for translations and rotations respectively. The resulting samples form the ground truth distribution. As a measure of the quality of the distribution estimated by a method we compute the Kullback-Leibler (KL) divergence {\blue and overlapping coefficient ($OVL$)}~\cite{ovl} between the obtained distribution and the ground truth distribution. {\blue 
A value of $0$ and $1$ for $OVL$ indicates no-overlap and full overlap between the two probability distributions respectively. 
}

In all experiments, Stein ICP uses Adam \cite{kingma2015} as the optimizer with a step size of $0.03$ and batch size of $150$ for the RGB-D data and a step size of $0.01$ and batch size of $300$ for the Challenging dataset~\cite{laserdataPomerleau2012ChallengingDS}. The bandwidth parameter required by SVGD is chosen using the median heuristic as described in \cite{steinNIPS2016_6338}. Furthermore, uniform priors are assumed as no prior knowledge is available. Bayesian ICP generates $1000$ samples and uses a step size of $0.008$. Cov-3D and Closed-form ICP use point-to-plane SGD-ICP estimated pose to compute the covariance matrix using the default parameter setting of the author's implementation \cite{ral}.
\subsection{Particle count}
 \begin{figure}[bt]
    \centering

%
    \begin{subfigure}{0.48\textwidth}
	    \includegraphics[page=1,width=0.98\textwidth]{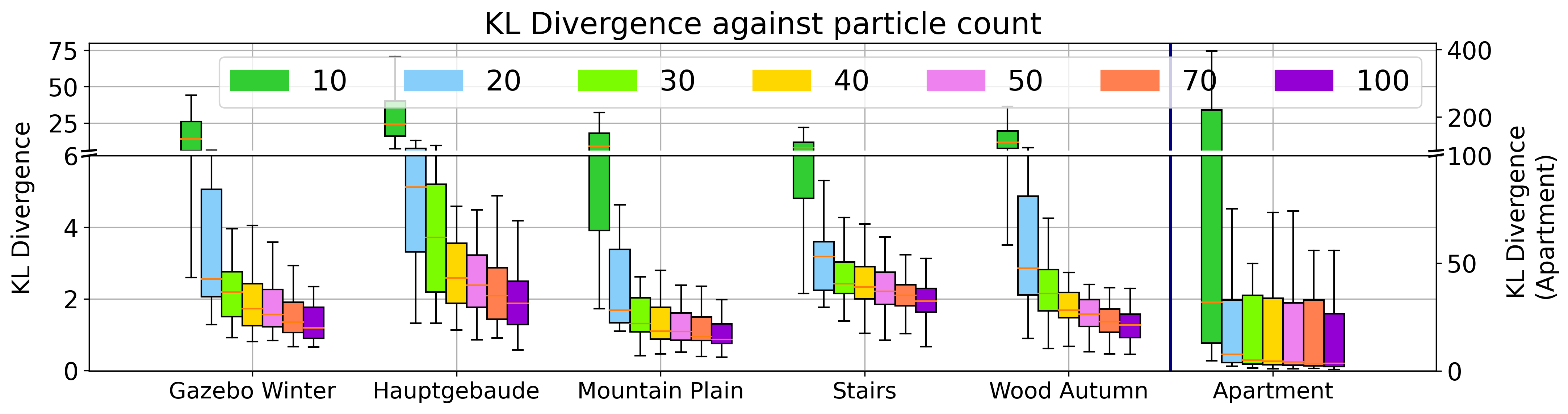}
	    
    \end{subfigure}
    
    \setlength{\belowcaptionskip}{-12pt}
    \caption{{\blue KL Divergence against particle count. Overall, the best performance is achieved with $100$ particles. Note however that there is little improvement in KL divergence between $40$ and $100$ particles.}}
    \label{fig:particlesize}	
\end{figure}

We begin by evaluating the effect the number of particles has on the pose distribution quality on the Challenging dataset. Figure \ref{fig:particlesize} shows the KL divergence between the Stein ICP and ground truth distribution using the point-to-point error metric as a function of the particle count. The right Y-axis is for the \emph{Apartment} sequence only due to the difference of the scale. Overall the KL divergence for both translations (top) and rotations (bottom) improves with a larger number of particles. However, the improvement in KL divergence between $40$ and $100$ particles is minor compared to the change from $10$ to $40$ particles. In the subsequent experiments we will use $100$ particles as this reliably achieves high quality results.
 
 \subsection{Distribution Quality Analysis}
 
\begin{figure}[bt]
    \centering
    \begin{subfigure}{0.48\textwidth}
    \hspace{1cm}
    \includegraphics[page=0.9,width=0.3\textwidth]{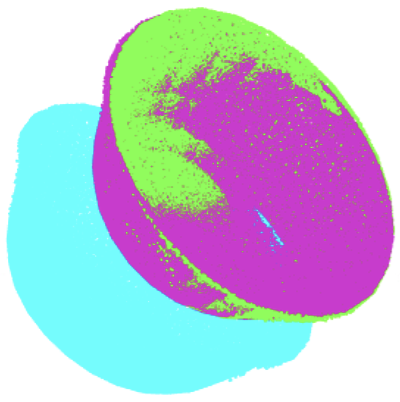}
    \hspace{1.2cm}
    \includegraphics[page=0.9,width=0.3\textwidth]{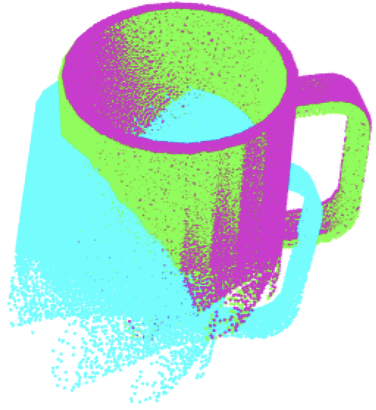}
    \end{subfigure}
    \vspace{-3mm}
    \setlength{\belowcaptionskip}{-15pt}
    \caption{Source (cyan), reference (magenta) and aligned cloud (green) of the \emph{bowl} (left) and the \emph{mug} (right). The \emph{bowl} is rotationally symmetric around yaw, while handle of the \emph{mug} constrains the yaw estimate.}
\label{fig:geom_clouds}	
\end{figure}

\begin{figure}[bt]
    \centering

    \begin{subfigure}{0.48\textwidth}
	    \includegraphics[page=1,width=0.49\textwidth]{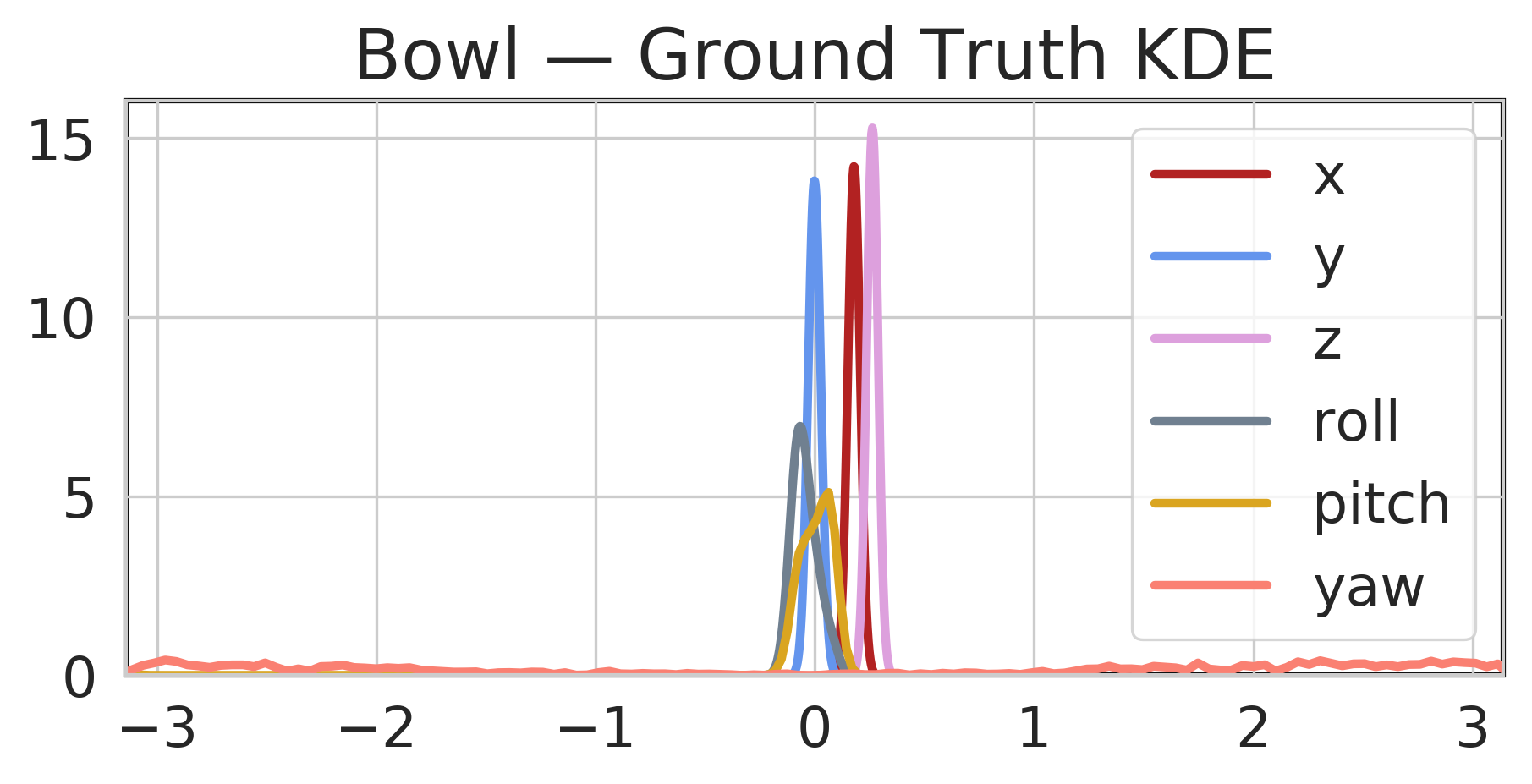}
	   \hfill
	    \includegraphics[width=0.49\textwidth]{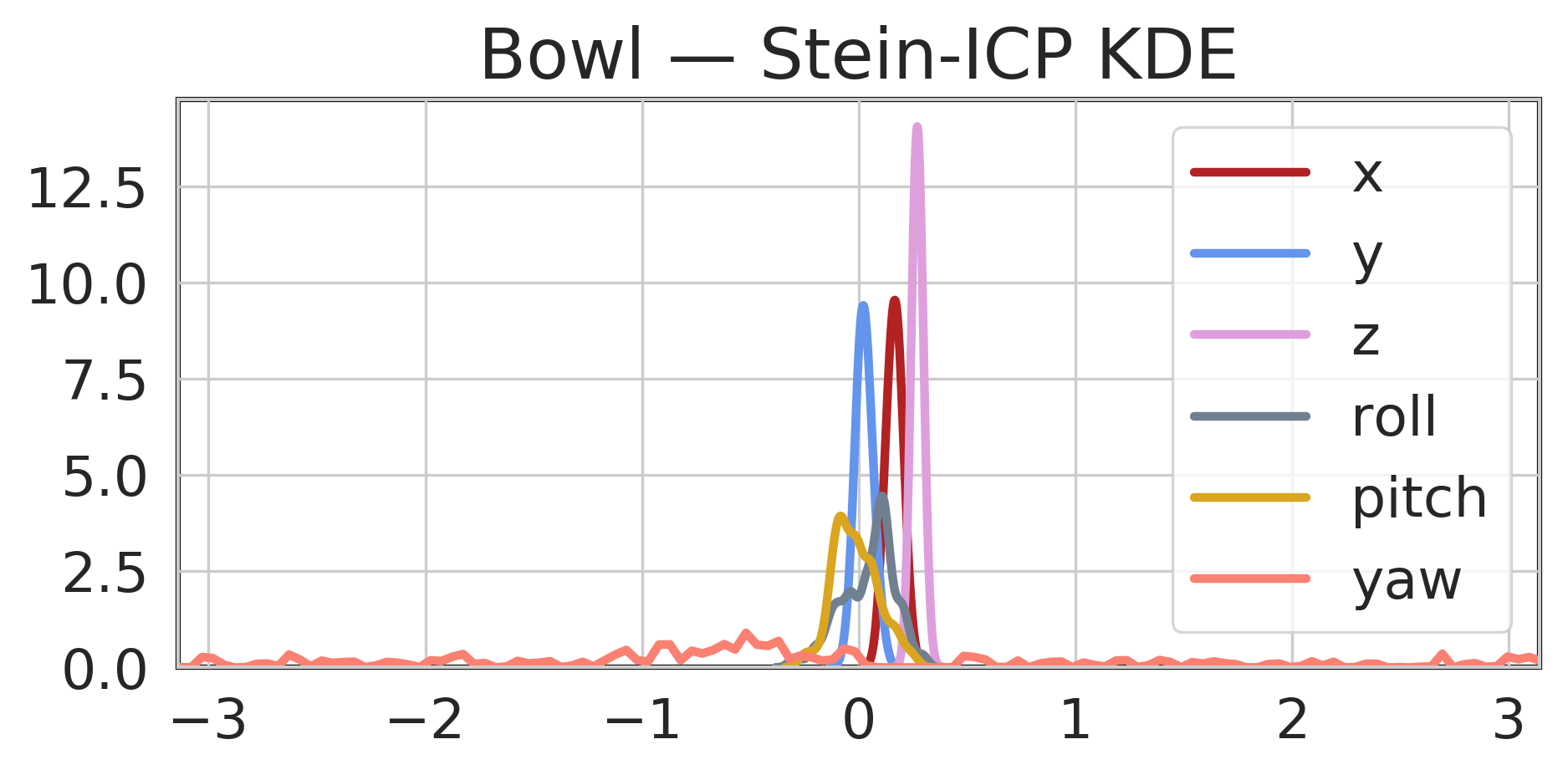}
        
    \end{subfigure}
   
    \begin{subfigure}{0.48\textwidth}
    	\includegraphics[page=1,width=0.49\textwidth]{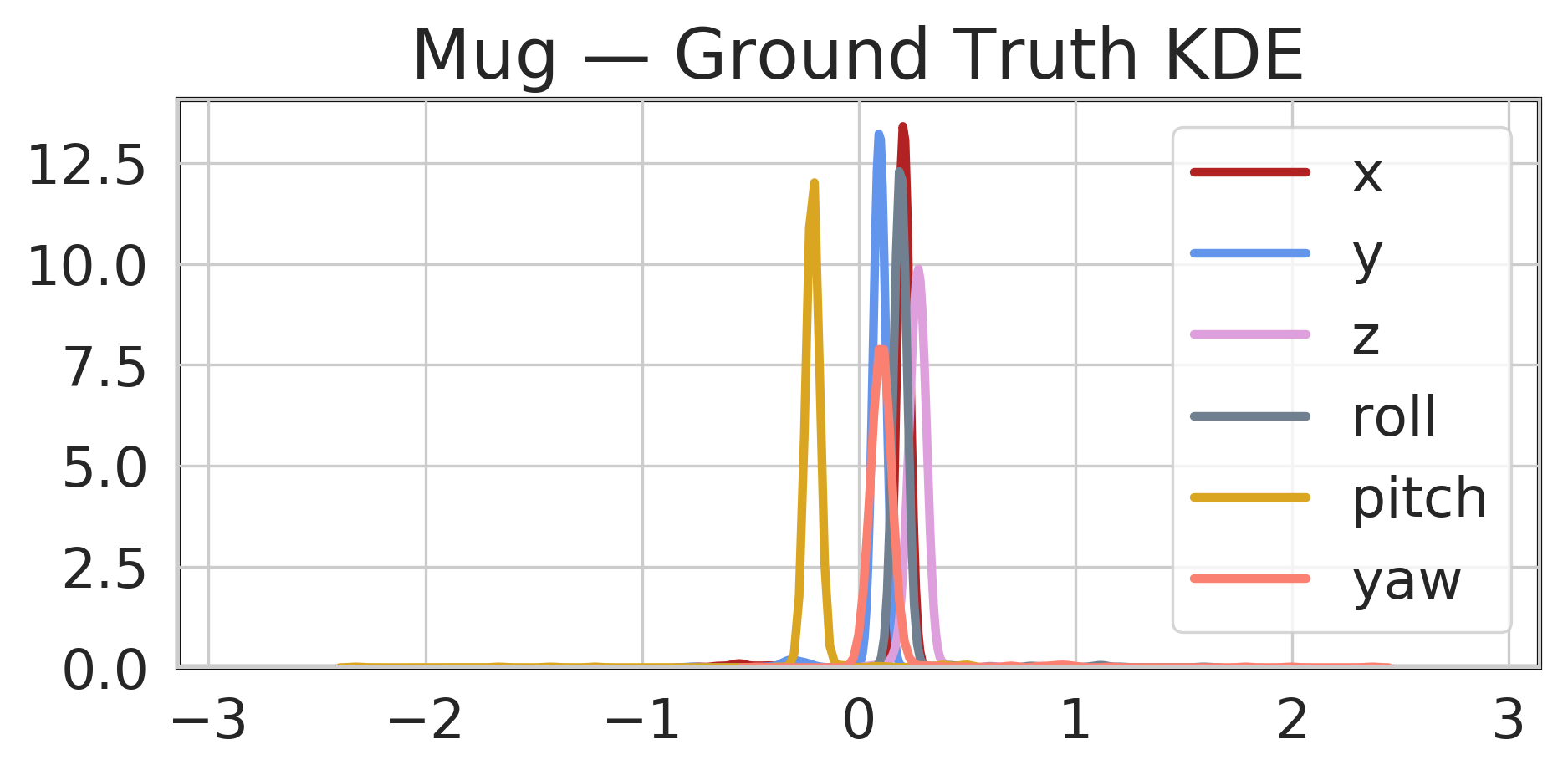}
    	\hfill
    	\includegraphics[width=0.49\textwidth]{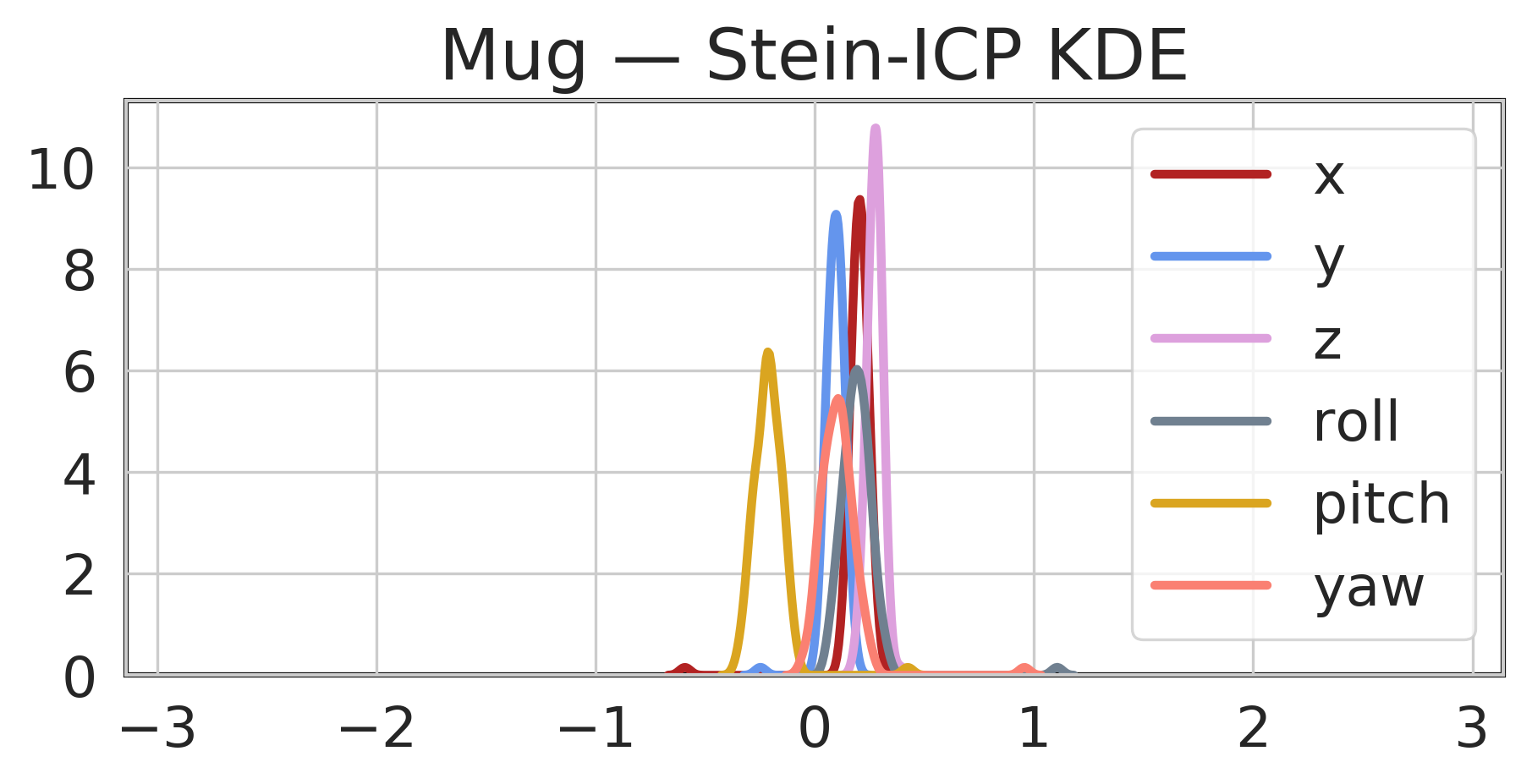}
      
    \end{subfigure}
    
    \setlength{\belowcaptionskip}{-16pt}
    \caption{Visualization of the pose distribution estimated by a kernel density estimator (KDE) for the ground truth and Stein ICP samples. 
    The broad yaw distribution correctly captures the rotational symmetry of the round \emph{bowl} (top). This is in contrast to the peaked yaw distribution of the \emph{mug} (bottom) which has no symmetries.}
   
    \label{fig:kdes}	
\end{figure}

We start with a qualitative evaluation of the distribution estimated by Stein ICP on two examples from the RGB-D dataset. In Figure \ref{fig:geom_clouds} we show the source (cyan), reference (magenta), and aligned (green) clouds for a \emph{bowl} (left) and a \emph{mug} (right). The kernel density estimates (KDE) of both the ground truth samples (left) and Stein ICP samples (right) are shown in Figure \ref{fig:kdes}. We can see how for the \emph{bowl} (top), which has a rotational symmetry around yaw, the estimates are peaked with the exception of yaw which spans the full $[-\pi, \pi]$ range, both in the ground truth distribution and Stein ICP. By contrast the distributions for the \emph{mug} (bottom) are peaked for all parameters. As the handle of the \emph{mug} clearly identifies the single correct solution, this uni-modal distribution is what we would expect. These results demonstrate that the distributions recovered by Stein ICP match  the ground truth distributions and that both distributions agree with our intuitions about object alignment ambiguity and uniqueness.

\begin{table}[bt]
	\begin{center}
		
		\caption{ Quality Comparison of the Stein ICP against other methods using median KL Divergence and {\blue overlapping coefficient (OVL)} with a $10^{th}$ and a $90^{th}$ quantile  written in the brackets.}
		\label{tab:kl_mahala}

	    \begin{tabular}{l l>{\blue}c>{\blue}c}
    		\toprule   
    		  
    	    \multirow{1}{*}{\raisebox{-\heavyrulewidth}{Sequence}} & \multirow{1}{*}{\raisebox{-\heavyrulewidth}{Method}} & \multicolumn{1}{c}{KL Div.} &
    	   \multicolumn{1}{c}{ $OVL$} \\
           

\cmidrule{1-4}

\multirow{5}{*}{Apartment}
& Ours  & $5.7 \;[ 1.2 , 4\mathrm{e}^{2} \;] $ & 
$0.7 \;[ 0.4 , 0.9 \;] $\\\
& {\blue Bayesian-multi}& $4.2 \;[ 0.8 , 1\mathrm{e}^{2} \;] $ & 
$0.7 \;[ 0.3 , 0.9 \;] $\\
& Bayesian ICP &$4.6 \;[ 1.6 , 2\mathrm{e}^{2} \;] $ & 
$0.7 \;[ 0.4 , 0.8 \;] $\\
& COV-3D ICP       & $2\mathrm{e}^{2} \;[ 17.2 , 2\mathrm{e}^{4} \;] $ &$0.3 \;[ 0.1 , 0.5 \;] $\\
& Closed-form ICP & $1\mathrm{e}^{4} \;[ 8\mathrm{e}^{2} , 9\mathrm{e}^{6} \;] $ & $0.0 \;[ 0.0 , 0.0 \;] $

\\ \vspace{-0.75em} \\
\multirow{5}{*}{Gazebo Winter}
& Ours                         & $1.1 \;[ 0.5 , 3\mathrm{e}^{2} \;] $ & $0.9 \;[ 0.4 , 0.9 \;] $\\
& {\blue Bayesian-multi}& $2.9 \;[ 1.0 , 4\mathrm{e}^{2} \;] $ &
$0.8 \;[ 0.4 , 0.9 \;] $\\
& Bayesian ICP &$3.0 \;[ 1.2 , 3\mathrm{e}^{2} \;] $& 
$0.8 \;[ 0.4 , 0.9 \;] $\\

& COV-3D ICP       & $42.4 \;[ 7.7 , 1\mathrm{e}^{3} \;] $ & 
$0.3 \;[ 0.1 , 0.4 \;] $\\
& Closed-form ICP & $1\mathrm{e}^{5} \;[ 4\mathrm{e}^{4} , 1\mathrm{e}^{7} \;] $ &
$0.0 \;[ 0.0 , 0.0 \;] $

\\ \vspace{-0.75em} \\
\multirow{5}{*}{Hauptgebaude}
& Ours                         & $0.6 \;[ 0.3 , 17.6 \;] $ &
$0.9 \;[ 0.8 , 0.9 \;] $\\
&{\blue Bayesian-multi}& $1.9 \;[ 1.3 , 13.2 \;] $ &
$0.8 \;[ 0.7 , 0.8 \;] $\\
& Bayesian ICP & $1.8 \;[ 1.3 , 15.7 \;] $ &
$0.8 \;[ 0.7 , 0.8 \;] $\\

& COV-3D ICP       & $2\mathrm{e}^{2} \;[ 55.5 , 8\mathrm{e}^{2} \;] $& $0.2 \;[ 0.1 , 0.3 \;] $\\
& Closed-form ICP & $4\mathrm{e}^{4} \;[ 1\mathrm{e}^{4} , 1\mathrm{e}^{5} \;] $ & $0.0 \;[ 0.0 , 0.0 \;] $
 
\\ \vspace{-0.75em} \\

\multirow{5}{*}{Mountain Plain}
& Ours  & $2.1 \;[ 1.6 , 12.7 \;] $ & 
$0.7 \;[ 0.6 , 0.9 \;] $\\
& {\blue Bayesian-multi}& $62.0 \;[ 1.9 , 5\mathrm{e}^{2} \;] $ 
& 
$0.5 \;[ 0.3 , 0.8 \;] $\\
& Bayesian ICP & $88.9 \;[ 4.3 , 3\mathrm{e}^{2} \;] $& 
$0.5 \;[ 0.3 , 0.7 \;] $\\

& COV-3D ICP       & $16.1 \;[ 6.4 , 69.4 \;] $ & $0.3 \;[ 0.2 , 0.5 \;] $\\
& Closed-form ICP & $4\mathrm{e}^{4} \;[ 1\mathrm{e}^{4} , 1\mathrm{e}^{5} \;] $ & $0.0 \;[ 0.0 , 0.0 \;] $

\\ \vspace{-0.75em} \\

\multirow{5}{*}{Stairs}
& Ours                         & $2.6 \;[ 0.9 , 3.7 \;] $ &
$0.7 \;[ 0.5 , 0.8 \;] $\\
& {\blue Bayesian-multi}& $1.9 \;[ 1.0 , 5.7 \;] $ & 
$0.8 \;[ 0.5 , 0.9 \;] $\\
& Bayesian ICP & $2.1 \;[ 0.9 , 4.9 \;] $&
$0.7 \;[ 0.5 , 0.9 \;] $\\
& COV-3D ICP       & $16.1 \;[ 11.3 , 1\mathrm{e}^{2} \;] $ & 
$0.3 \;[ 0.2 , 0.4 \;] $\\
& Closed-form ICP & $4\mathrm{e}^{3} \;[ 6\mathrm{e}^{2} , 8\mathrm{e}^{4} \;] $ & $0.0 \;[ 0.0 , 0.0 \;] $

\\ \vspace{-0.75em} \\

\multirow{5}{*}{Wood Autumn}
& Ours                         & $0.6 \;[ 0.4 , 1.7 \;] $ & 
$0.9 \;[ 0.8 , 0.9 \;] $\\
& {\blue Bayesian-multi} & $1.2 \;[ 0.8 , 2.5 \;] $ &
$0.9 \;[ 0.8 , 0.9 \;] $\\
& Bayesian ICP & $1.2 \;[ 0.7 , 3.2 \;] $& 
$0.9 \;[ 0.6 , 0.9 \;] $\\
& COV-3D ICP       & $8.1 \;[ 7.1 , 9.4 \;] $ & 
$0.4 \;[ 0.3 , 0.5 \;] $\\
& Closed-form ICP & $4\mathrm{e}^{4} \;[ 2\mathrm{e}^{4} , 1\mathrm{e}^{5} \;] $ & $0.0 \;[ 0.0 , 0.0 \;] $

\\ \vspace{-0.75em} \\

		    \bottomrule 
	    \end{tabular}
	\end{center}
	\vspace{-9mm}
\end{table}

Next, we perform a quantitative comparison between our method and the state-of-the-art probabilistic ICP methods namely \emph{Bayesian ICP} \cite{Maken2020EstimatingMU}, {\blue a multi-chain Bayesian ICP--denoted by Bayesian-multi--this is the result of running $10$ chains of Bayesian ICP and compounding the results}, \emph{COV-3D ICP} \cite{ral}, and a popular benchmark \emph{Closed-form ICP} \cite{Censi2007} on the Challenging dataset. The results are summarised in Table \ref{tab:kl_mahala} which shows the median of the KL divergence {\blue and $OVL$} as well as the $10^{th}$ and $90^{th}$ quantile in the brackets. The divergence {\blue and $OVL$ are} computed over all consecutive pairs in each sequence 
with all methods using the point-to-plane error metric.

From those numbers we can see that our method outperforms both \emph{COV-3D} and \emph{Closed-form} ICP on all of the sequences. 
The large values of KL divergence {\blue and small values of OVL} for \emph{COV-3D} and \emph{Closed-form} ICP indicate that these methods fail to provide correct uncertainty estimates. The \emph{Closed-form} ICP method only provides reliable estimates in situations where uncertainty arises solely due to sensor noise \cite{ral}, which is not the case in many of the scenes in the Challenging dataset. The \emph{COV-3D} method relies on an accurate initial pose uncertainty estimation to provide a correct pose covariance. This causes the algorithm to 
{\blue under- or over-estimate the pose covariance} when the initial covariance estimate is either inexact or not available~\cite{ral}.

When comparing Stein ICP with \emph{Bayesian ICP} {\blue and \emph{Bayesian-multi}} we observe that in most scenes {\blue Stein ICP performs better. The only exception is \emph{Apartment} where \emph{Bayesian ICP} performs better.} As the KL divergence is computed based on the covariance of the samples, the fact that Stein ICP only produces $100$ samples while Bayesian ICP produces $1000$ samples can have an impact on the KL divergence. However, overall, both methods produce results of comparable quality. 
 {\blue 
 Similar results can be observed for the $OVL$ measure, where Stein ICP performs better overall.}

\subsection{Odometry}

In this section we use Stein ICP to compute odometry estimates by propagating the frame-to-frame expectations of the ICP transformation distributions to obtain a global pose estimate. These results are compared with the odometry estimate obtained by Bayesian ICP. As our method provides samples of the transformation parameters and can capture multi-modal distributions, the overall expectation and covariance may not represent the underlying correct pose.

\begin{figure}[bt]
    \centering
\begin{subfigure}{0.48\textwidth}
		\includegraphics[page=1,width=0.48\textwidth]{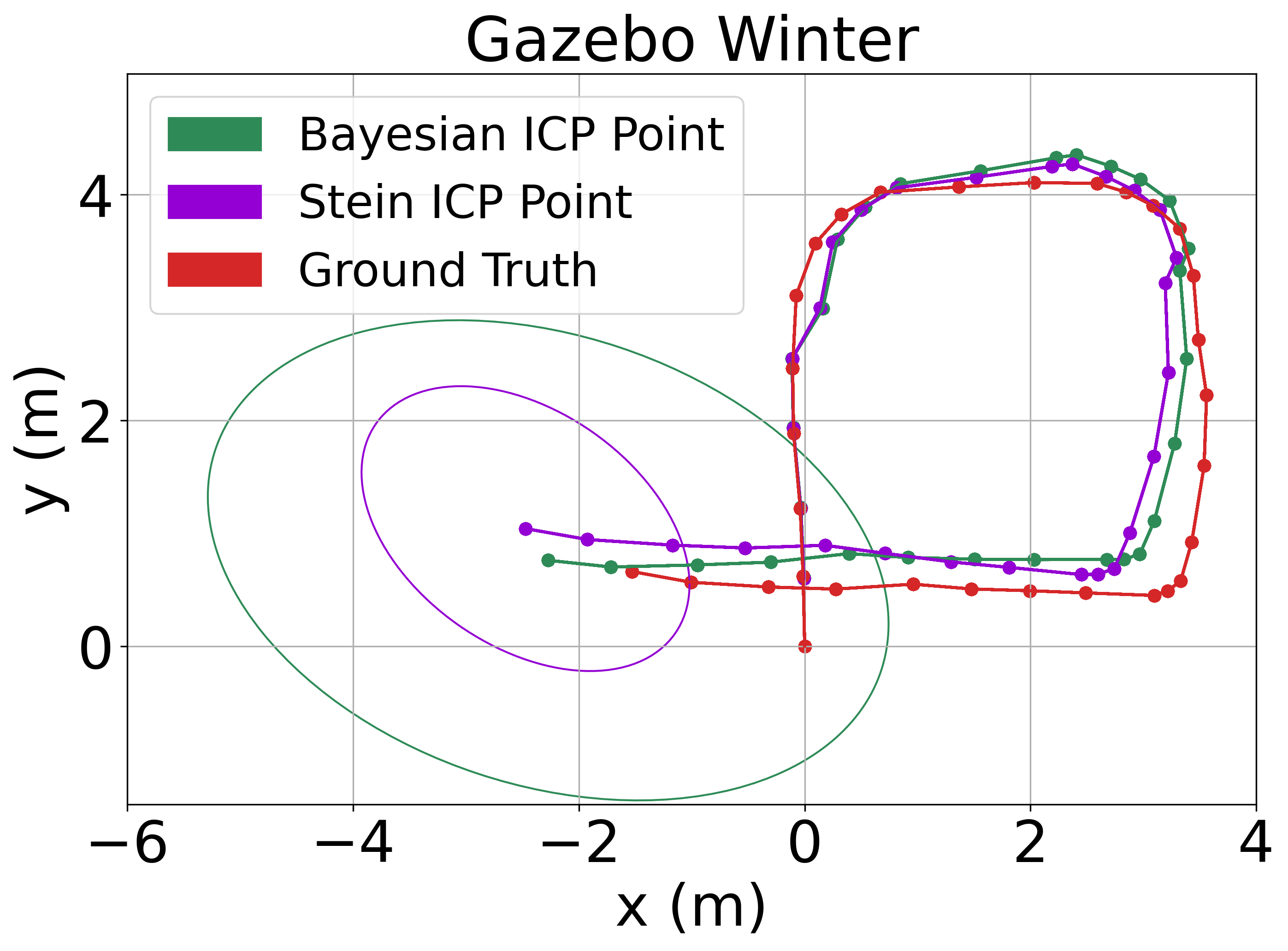}
    	\hfill
    \includegraphics[page=1,width=0.48\textwidth,height=3 cm]{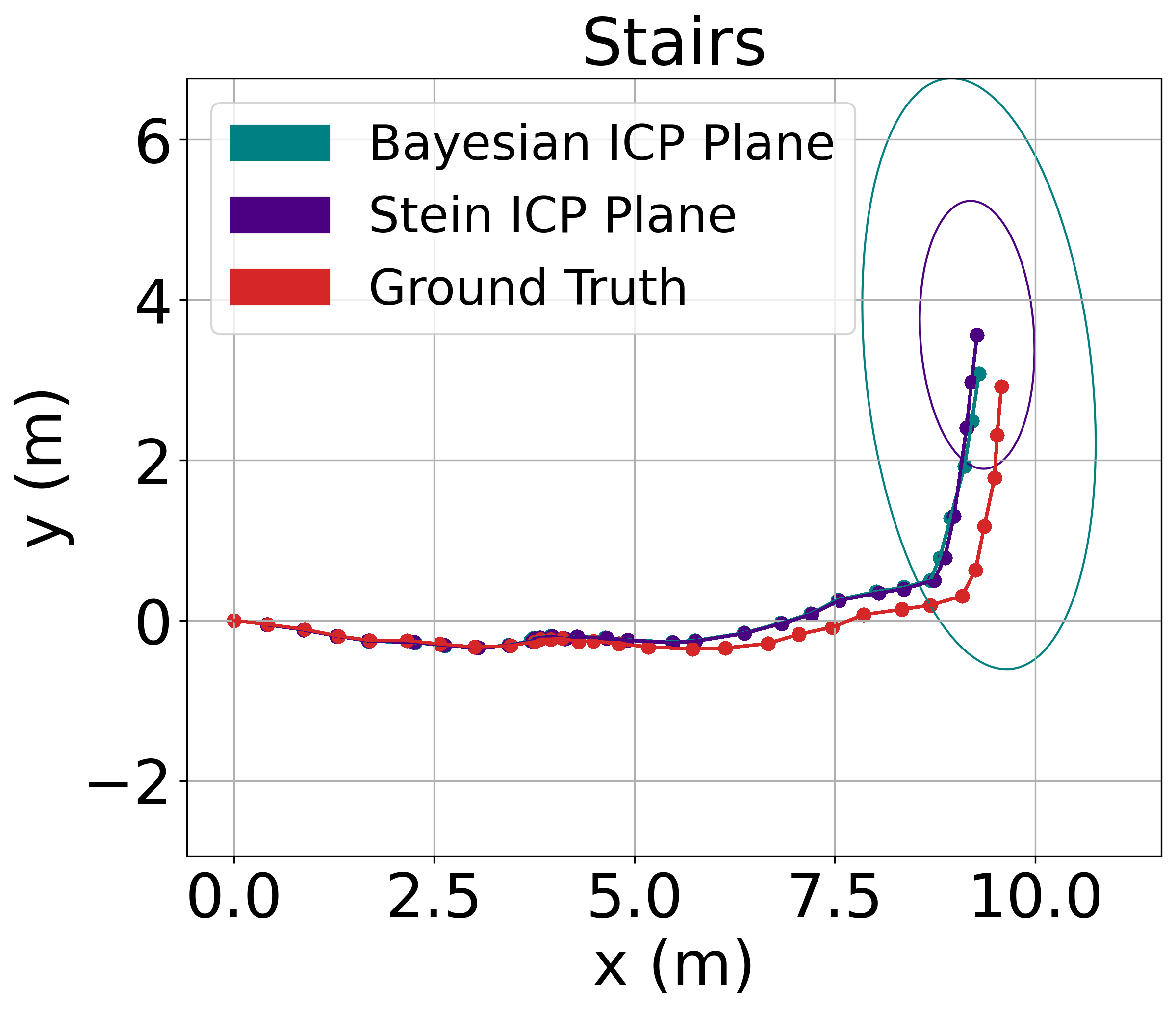}
    \end{subfigure}
    
    \begin{subfigure}{0.48\textwidth}

	   \includegraphics[page=1,width=0.98\textwidth]{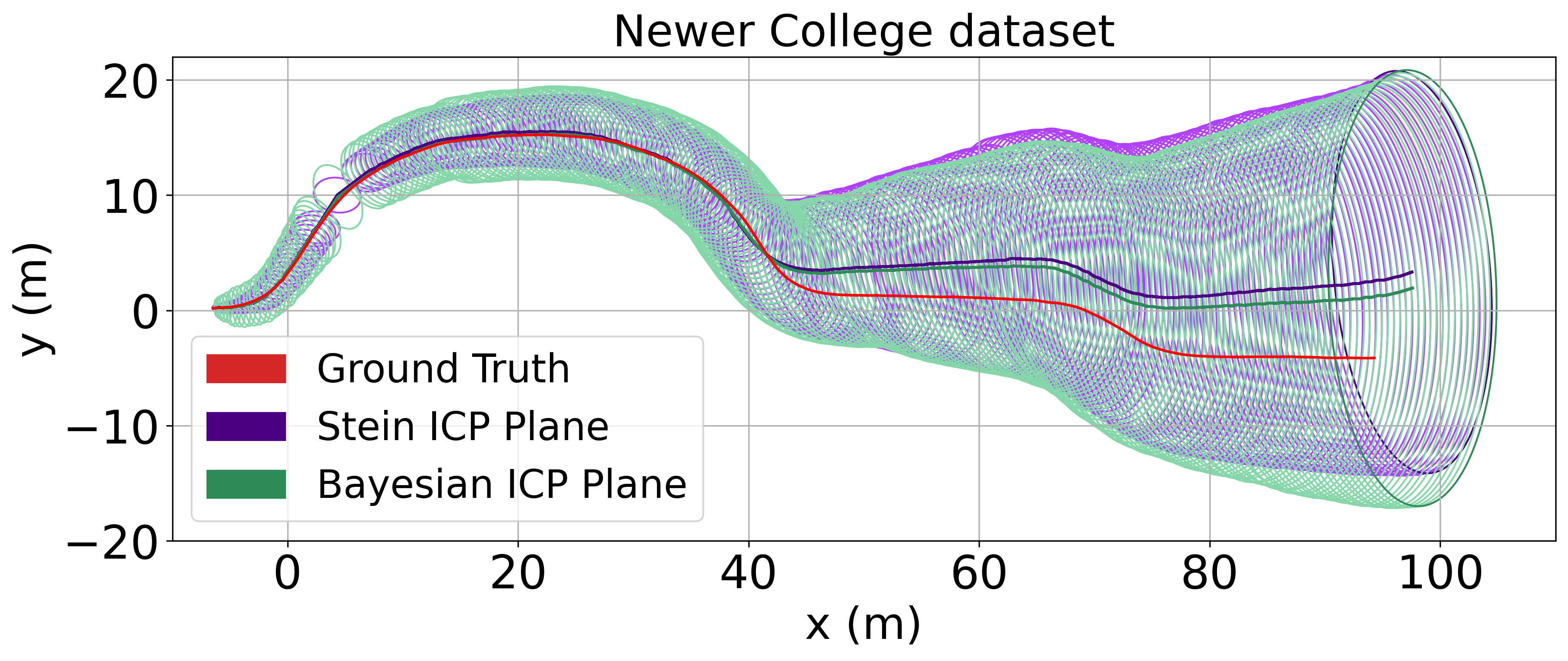}
    \end{subfigure}
    \vspace{-3mm}
    \setlength{\belowcaptionskip}{-11pt}
    \caption{Stein ICP and Bayesian ICP odometry overlaid on the ground truth trajectory for \emph{Gazebo Winter}, \emph{Stairs}, \blue {and Newer College dataset~\cite{newcollege}}. The ellipses represent the $95\%$ confidence set of the compound covariance at the last pose of odometry for each method {\blue for sequences and along the entire trajectory for~\cite{newcollege}}. 
    }
  
    \label{fig:Odometry}	
\end{figure}

For a sequence of $l$ scans the individual ICP pose estimates are combined as $T_E=T^1_0 T^2_1 ....  T^l_{l-1}$. Where $T^l_{l-1}$ is a $4 \times 4$ homogeneous transformation matrix that captures the mean rotation and translation between scan $l$ and $l-1$. A $4^{th}$ order approximation \cite{barfoot6727494} is used to combine the individual scan-to-scan covariances. The resulting trajectories for \emph{Gazebo Winter}, \emph{Stairs}, {\blue and Newer College dataset~\cite{newcollege}} are overlaid onto the ground plane to visualize the estimated trajectories in Figure \ref{fig:Odometry}. The lines display the trajectories and the ellipses show the $3-sigma$ ($95\%$) confidence sets for uncertainty estimation of Stein ICP and Bayesian ICP. 

The plots in Figure \ref{fig:Odometry} show that the mean trajectory obtained from Stein ICP and Bayesian ICP estimates are similar and track the ground truth trajectory well. The covariance ellipses estimated using Bayesian ICP are larger and less certain than those of Stein ICP. As the odometry estimates are obtained by accumulating transformations 
small differences in errors can add up over the trajectory, amplifying the small differences between Stein ICP and Bayesian ICP reported in Table \ref{tab:kl_mahala}. Overall both methods provide good odometry trajectories and covariance estimates. The ability to estimate odometry trajectories including uncertainty is crucial in many robotic applications such as autonomous navigation and SLAM. More generally any method reliant on pose estimation can use the uncertainty provided by these methods for decision making.

\subsection{Impact of Distribution Estimation on Mean Pose Estimation Quality}

  \begin{figure}[bt]
    \centering

    \begin{subfigure}{0.48\textwidth}
	    \includegraphics[page=1,width=0.96\textwidth]{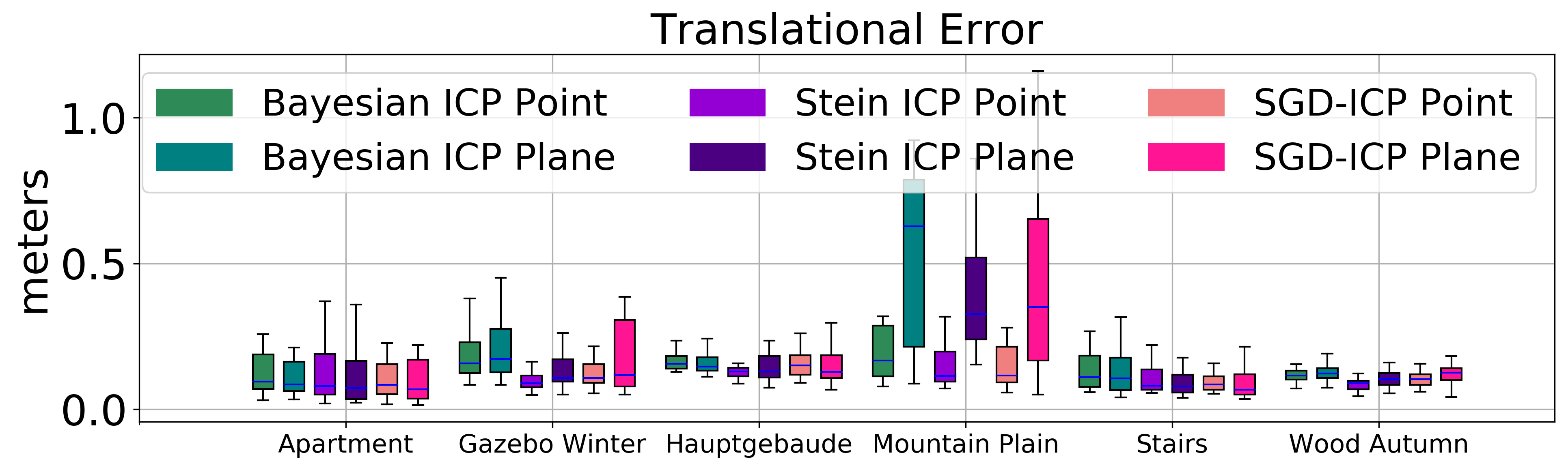}
	    
    \end{subfigure}

    \begin{subfigure}{0.48\textwidth}
    	\includegraphics[page=1,width=0.96\textwidth]{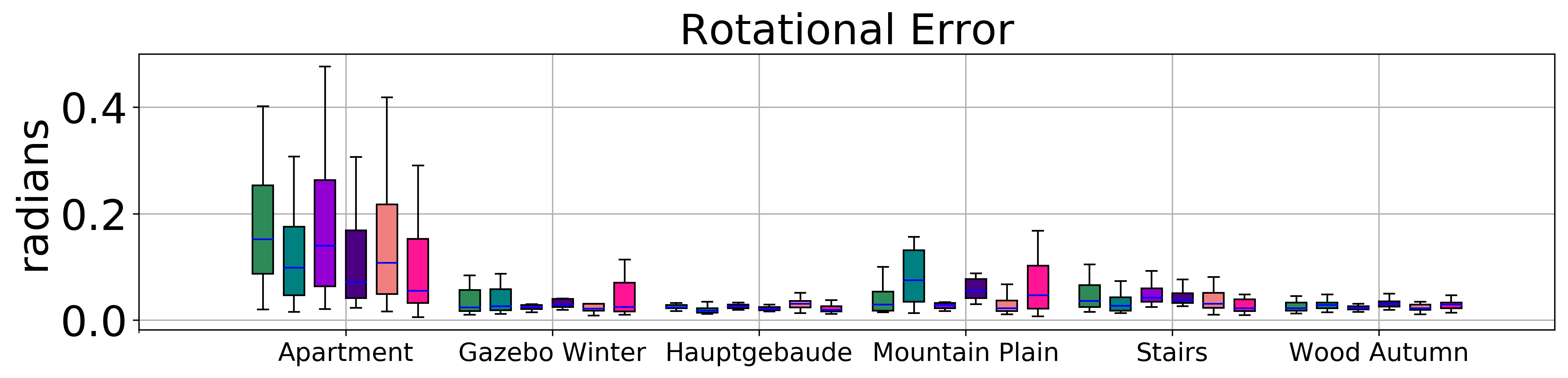}
    	
    \end{subfigure}
    \setlength{\belowcaptionskip}{-15pt}
    \caption{Translational (top) and rotational (bottom) error distributions over entire trajectories of Stein ICP in comparison to Bayesian ICP and SGD-ICP method on the Challenging dataset. Overall, Stein ICP  gives equivalent performance to that of SGD-ICP indicating that Stein ICP does not suffer from a reduction in quality despite providing uncertainty estimates.}
   
    \label{fig:traj_err}	
\end{figure}

Estimating the full distribution over pose parameters is useful. However, if a single solution exists what, if any, is the price in pose estimation accuracy being paid in comparison to a standard point estimation method? To answer this question we examine the trajectory estimation errors for Stein ICP, Bayesian ICP, and SGD-ICP all using both the point-to-point and the point-to-plane error metric. To measure the local accuracy of the trajectory estimation we compute the \emph{relative pose error} \cite{slambencSturm2012} (RPE) for the SGD-ICP solution and each sample of the distributions produced by Stein ICP and Bayesian ICP. The RPE is defined as follows:
\begin{equation}
    E = {(\hat T_i^{-1} \hat T_{i+\Delta})}^{-1}( T_i^{-1} T_{i+\Delta}),
    \label{RPE}
\end{equation}
where $(\hat T_i^{-1} \hat T_{i+\Delta})$ is the ground truth transformation and $( T_i^{-1} T_{i+\Delta})$ is the ICP transformation between scan $i$ and $i+\Delta$, with $\Delta \ =1$ here.

The translational error is computed using the Euclidean distance of the translation components of $E$ while the rotational errors are computed using absolute angular differences between the orientation of the ground truth and the estimated pose. For sample based methods we obtain a single error by averaging together the per sample errors. In Figure \ref{fig:traj_err} we show the translational (top) and rotational (bottom) error distributions over all the consecutive scan pairs of each used sequence independently.

From this we can see that the estimates of the distribution based methods are not any worse than those of a point estimation method. The choice of error metric has a bigger impact on the results than whether or not a distribution is estimated. This means that we do no give up any accuracy as far as transformation estimation goes with Stein ICP.

\subsection{Runtime}

\begin{figure}[bt]
    \centering
	    \includegraphics[width=0.49\textwidth]{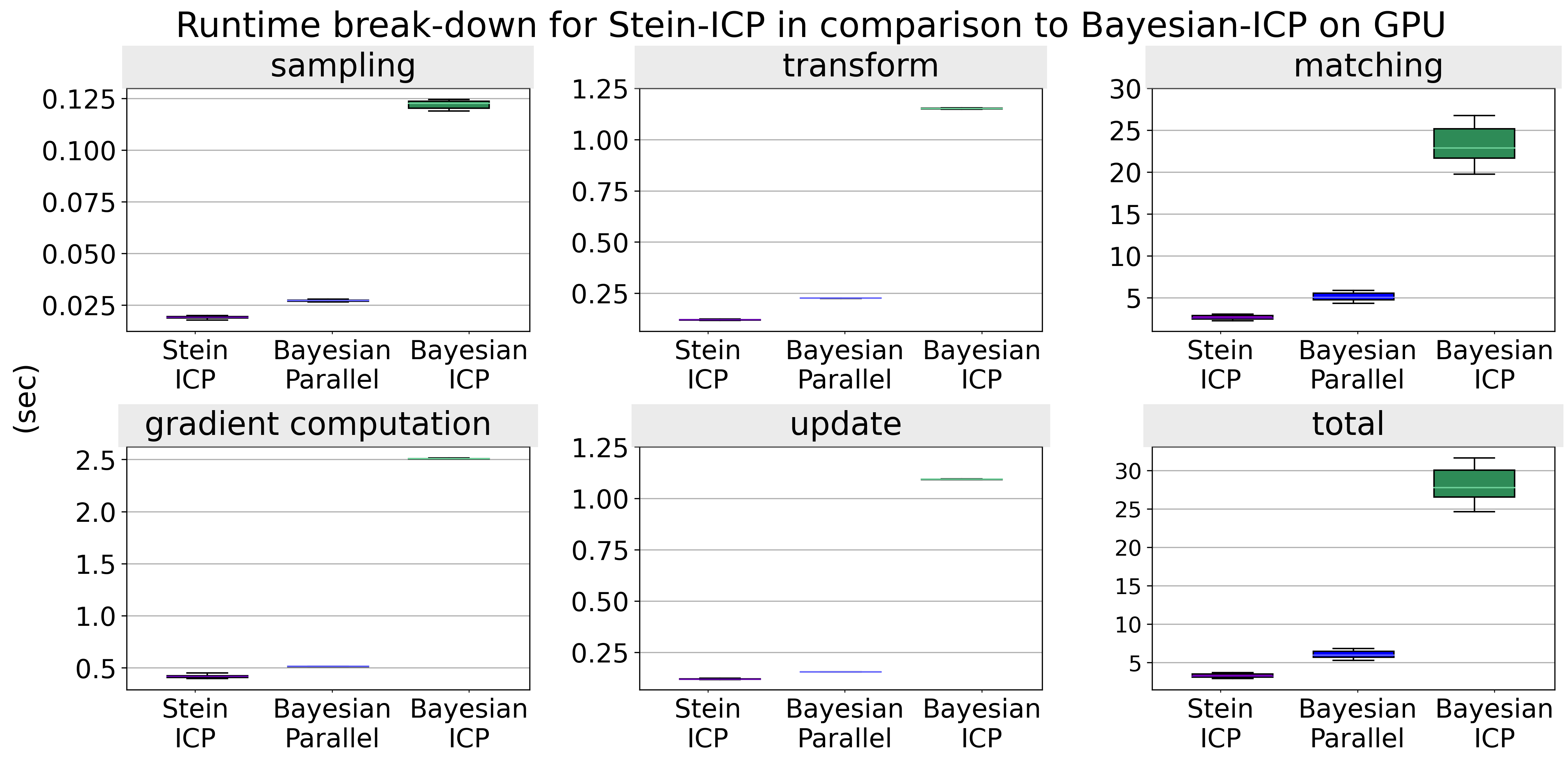}
    
    \vspace{-2mm}
    \setlength{\belowcaptionskip}{-15pt}
    \caption {{\blue Break-down of the runtime (sec) 
    on the GPU. The times correspond to mini-batch sampling (top left), transforming point cloud with the updated transformation time (top middle), matching time (top right), gradient time (bottom right), and update time (bottom middle). Bottom right shows the total time. For Stein ICP, the gradients computational time includes the kernel estimation and repulsive force. These results demonstrate that Stein ICP is more efficient than both versions of Bayesian ICP.}} 
    \label{fig:run-time}	
\end{figure}

Finally, we compare the run-time of Stein ICP, Bayesian ICP {\blue and Bayesian-multi}. Traditionally, obtaining uncertainty estimates for ICP methods was computationally expensive, especially if accurate but slow Markov Chain Monte Carlo based methods were used. Runtime efficiency is one of the main advantages of Stein ICP due to its parallelism and leverage of GPUs. In the following, we present a break down of the runtime by the individual components of Stein ICP compared to Bayesian ICP. To this end, we use the \emph{Gazebo Winter} dataset and run Stein ICP for $100$ iterations using $100$ particles. In Bayesian ICP, each iteration provides us with a sample. However we need to run the method over a certain number of iterations before the Markov Chain becomes stable; a process known as burn-in. To this end, we run Bayesian ICP {\blue for $1100$ iterations to collect $1000$ sequential samples after discarding initial $100$ burn-in samples. We also run $10$ parallel chains of Bayesian-multi where each chain collects $100$ samples 
with a burn-in period of $100$ iterations} to produce a distribution of equivalent quality. All computations are performed on a desktop computer with an NVidia Titan-V GPU.

Figure \ref{fig:run-time} presents the runtime of the various components along the Y-Axis. 
Even though Stein ICP is run with $100$ particles, which is roughly equivalent to running $100$ independent SGD-ICP instances, Stein ICP takes $\frac{1}{8^{th}}$ of the time Bayesian ICP requires for each ICP component. This is because Stein ICP's particles can trivially be parallelized on a GPU. In contrast, Bayesian ICP produces samples under the Markov assumption, \ie every sample depends on the previous one, therefore it cannot reliably exploit GPU parallelism to generate samples.
{\blue Even parallel Bayesian ICP (if run for efficiency) consumes relatively more time. This is due to the burn-in period necessary for each parallel chain of Bayesian ICP to ensure convergence to the true distribution}. Stein ICP ability to perform computations in parallel on a per particle basis means that the GPU implementation can do more work for $100$ particles in roughly the same time as required by a single particle. This results in over 5 times speedup compared to Bayesian ICP. 
{\blue Both Cov-3D and Closed-form methods require at least solving a point-to-plane ICP first ($2.62 \pm 0.19$ sec on the GPU) in addition to the time required to compute the covariances. In addition, Cov-3D requires $12$ registration steps for the unscented transform.} 
\section{Conclusion}
In this paper we devised Stein ICP, a probabilistic generalization of the popular ICP that provides accurate posterior pose distributions. Stein ICP exploits the stochastic optimization nature of SGD-ICP combining it with the recent Stein variational gradient descent (SVGD) algorithm to approximate posterior distributions of transformation parameters between two point clouds using particles. The independent nature of these particles makes Stein ICP suitable for GPU parallelization. Extensive experiments using both RGB-D as well as LiDAR data demonstrate the capability of our method in providing high-quality pose parameter distributions in only a few seconds of computation. In contrast to Markov chain Monte Carlo solutions to probabilistic point-cloud matching, Stein ICP propagates gradients of particles in parallel, connecting gradient descent on the objective function with a repulsive term that ensures particles are aligned and capture the tail of the underlying distribution. This opens up new possibilities to leverage calibrated uncertainty estimation in applications requiring robust point cloud matching such as simultaneous localization and mapping or 3D reconstruction. Another exciting direction on using the uncertainty provided by the method is for grasping unknown objects based on a list of known objects with associated grasping positions. The uncertainty can be used to indicate which of the known objects is a better match and which direction (with less uncertainty) to grasp the unknown object. 

\vspace{-6mm}
\section*{}
{\small
\bibliographystyle{plain}
\bibliography{stein_icp}
}

\end{document}